# Accelerating Post-Tornado Disaster Assessment Using Advanced Deep Learning Models

Robinson Umeike
Civil, Construction and Environmental Engineering & Computer Science.
The University of Alabama
Tuscaloosa, USA
crumeike@crimson.ua.edu

Thang Dao
Civil, Construction and Environmental Engineering.
The University of Alabama
Tuscaloosa, USA
tndao@eng.ua.edu

Shane Crawford
Civil, Construction and Environmental Engineering.
The University of Alabama
Tuscaloosa, USA
pscrawford@ua.edu

***Abstract*—Post-disaster assessments of buildings and infrastructure are crucial for both immediate recovery efforts and long-term resilience planning. This research introduces an innovative approach to automating post-disaster assessments through advanced deep learning models. Our proposed system employs state-of-the-art computer vision techniques—YOLOv11 and ResNet50—to rapidly analyze images and videos from disaster sites, extracting critical information about building characteristics, including damage level of structural components and the extent of damage. Our experimental results show promising performance, with ResNet50 achieving 90.28% accuracy and an inference time of 1529ms per image on multiclass damage classification. This study contributes to the field of disaster management by offering a scalable, efficient, and objective tool for post-disaster analysis, potentially capable of transforming how communities and authorities respond to and learn from catastrophic events.**

## I. Introduction

The increasing frequency and intensity of natural disasters, exacerbated by climate change, pose significant challenges to urban infrastructure and resilience of communities worldwide. A plain example is the December 2021 Quad-State Tornado (Arkansas, Missouri, Tennessee, and Kentucky), which carved a devastating 266.67 km path across the Midwest, causing $3.9 billion (2022 USD) in damages, at least 667 injuries, over 90 fatalities, and requiring several man-hours of post-disaster assessments [1]. This event, one of the longest-tracked tornado in U.S. history, highlighted critical gaps in our current disaster response capabilities.

In the aftermath of such tragic events, rapid and accurate assessment of building damage is crucial for effective emergency response, resource allocation, and long-term recovery planning [1]. Traditionally, post-disaster studies focusing on the performance of buildings and infrastructure have relied heavily on manual interpretation of vast quantities of data, including images and videos collected from disaster sites [2]. However, this approach requires trained non-experts or experts to visually inspect and categorize damage indicators, a process that can take several weeks for large-scale events [3], is cost-intensive, and potentially subject to biases that lead to inconsistencies in damage evaluation [4].

Moreover, effective recovery measures implies the swiftness of emergency response and rapid service restoration, as delays can exponentially increase disruption impacts. Rapid response is particularly crucial in scenarios that demand swift action, such as during aid delivery and evacuation missions. The success of these operations often depends on the pace of execution; lengthy delays can render these efforts futile.

Recent advances in artificial intelligence (AI), specifically deep learning and computer vision, are starting to have direct applicability in disaster management [5]. These techniques can be used to swiftly analyze video footage from widespread CCTV networks, satellites, and ground cameras to identify incidents as they unfold, assess infrastructure damage, detect weaknesses in response strategies, and track resource needs in real-time. Such capabilities allow communities to mount rolling responses that contain dangers and restore functionality of its critical infrastructures through continuously optimized response actions.

TABLE I. Summary attributes for multitask post-disaster AI assessment

| Attribute | Task Description | |
|---|---|---|
| | ***Damage Classification*** | ***Damage Detection*** |
| Dataset | 2013 Moore Tornado | 2021 Midwest Tornado (Mayfield) |
| Images Count (Annotations) | 2635 (2635) | 1776 (2816) |
| Class Taxonomy | undamaged building, roof damage, wall collapse, wall-&-roof damage, not-a-building | undamaged, slight, moderate, extensive, and complete |
| Model Architecture | ResNet50 | YOLOv11 |
| Data Proprocessing | Resizing, Rotation, Shear and Zoom, Horizontal flipping, Width, and height shifts. | Bounding boxes, Auto-orientation, Filter null images, Resizing, Rotation. |
| Data Augmentation | No | Yes |
| Data Split (Train:Val:Test) | 80:10:10 | 73:18:9 |
| Training Paramaters | Batch size: 64, Epochs: 5, Optimizer: Adam, Learning Rate: 0.0001 | Batch size: 8, Epochs: 30, Optimizer: SGD, Learning Rate: 0.01 |
| Evaluation Metrics | Learning Curves (Model Accuracy and Loss), Precision-Recall Curve, and Confusion Matrix | Training Losses, Precision, Recall, Mean Average Precision (mAP), |
| Model Accuracy | 90.28% | 60.83% |
| Inference Time on 1 NVIDIA GeForce RTX 4090 24G GPU | 1529ms | 3.0ms |

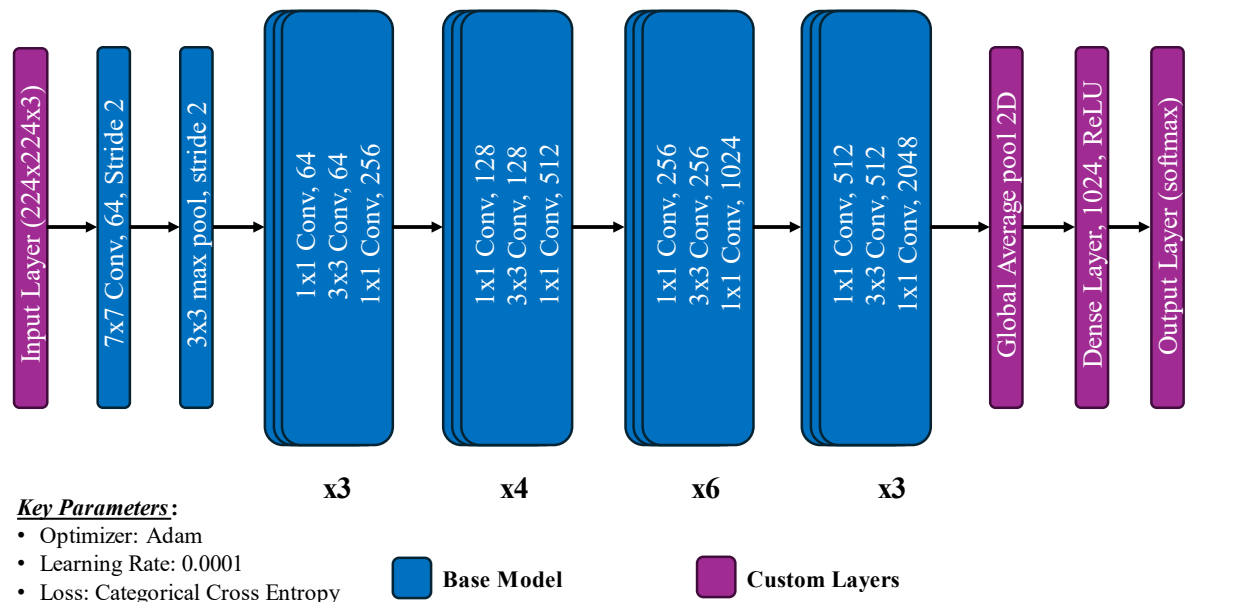

**Fig. 1.** Modified ResNet50 Architecture for Transfer Learning [4].

Although current AI-based approaches have shown great promise, with reported accuracies as high as 99.82% [6], they have primarily focused on earthquake, flood, or hurricane damage, leaving a significant gap in tornado-specific damage assessment[5]. The unique characteristics of tornado damage—including their narrow paths, extreme localized destruction, and complex damage patterns—present distinct challenges that distinguish it from other natural disasters.

This research presents an innovative automated disaster analysis framework for post-tornado damage assessment that leverages advanced deep learning methods for extracting critical information about structural composition, and damage states. The core of our approach utilizes two prominent deep learning architectures: YOLOv11 (You Only Look Once version 11) for object detection, classification, and localization of buildings within complex post-disaster scenes [7]; and ResNet50 (Residual Network with 50 layers) for fine-grained classification of building component damage detection [8].

Our approach aims to address the limitations of manual methods by providing a fast, consistent, and scalable solution for analyzing visual data from disaster sites (Table 1). Its key objectives are to:

i. Accelerate post-disaster damage assessment and relief effort using advanced deep learning techniques.

ii. Identify efficient techniques for minimizing resource utilization and errors proceeding from non-standardized manual post-disaster damage evaluation.

iii. Automate the data curation pipeline to help facilitate faster disaster data annotation during future post-disaster studies.

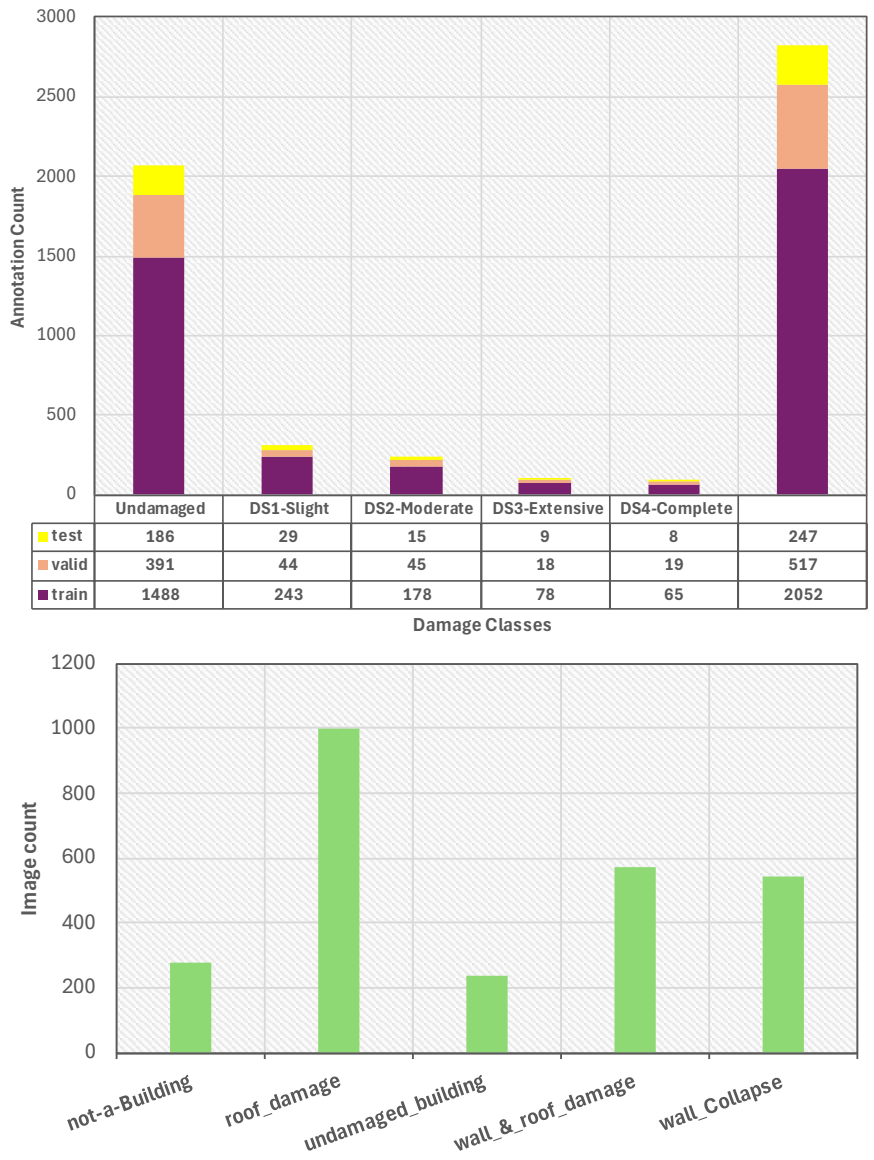

**Fig. 2.** Damage class distribution: **Top**: 2021 Midwest Tornado (Mayfield, KY), **Bottom**: 2013 Moore Tornado.

## II. Methods

### A. Data Curation and Preprocessing

The foundation of our research lies in a diverse dataset compiled from two significant tornado events (The 2021 Midwest Tornado and the 2013 Moore Tornado). For the former, we extracted images of damaged buildings from vehicle-mounted, 360° video data collected in Mayfield, Kentucky following the data collection and processing method in [2, 9]. The latter dataset comprises images taken using hand-held cameras in Moore, Oklahoma [10].

Our data processing pipeline employs a multi-faceted approach tailored to each dataset's characteristics. We utilized tagging tool for data annotation, preprocessing, and packaging of the 2021 Midwest Tornado data. This process involved outlining buildings with bounding boxes and categorizing damage into five classes using the INCORE damage state classification: undamaged, slight, moderate, extensive, and complete (top panel of Figure 2) [1]. In contrast, the 2013 Moore Tornado dataset underwent manual inspection to filter out low-quality images and noise. Damage in this dataset was categorized using a slightly different taxonomy: undamaged building, roof damage, wall collapse, wall-and-roof damage, and not-a-building as shown in the bottom panel of Figure 2.

*1) Damage Classification:* The images were organized into directories corresponding to their respective damage classification labels. To ensure consistent input dimensions for our deep learning model, all images were resized to 224x224 pixels with three color channels (RGB), resulting in an input shape of (224, 224, 3).

*2) Damage Detection:* First, we applied an auto-orientation procedure to correct EXIF metadata orientation inconsistencies in the source images. This technical correction ensures that all images are read with their pixels properly aligned with their intended viewing orientation, preventing inadvertent misalignment between image data and model training. Without this correction, images captured in different camera orientations might have their underlying pixel arrays misaligned, even if they appear correct to human viewers. Subsequently, all images were resized to a uniform dimension of 640x640 pixels. This resizing operation serves dual purposes. First, it ensures consistency in input size for the YOLOv11 model, and secondly, it optimizes the balance between preserving image detail and computational efficiency.

### B. Data Augmentation

*1) Damage Classification:* Data augmentation methods were employed to boost the robustness of our model and mitigate overfitting. These techniques were applied dynamically during the training process through a custom data generator. The generator applied rotations of up to 40 degrees, width and height shifts up to 20% of the image dimensions, as well as shear and zoom transformations within the same range. Horizontal flipping was also incorporated to increase data diversity. Finally, any newly created pixels resulting from these transformations were filled using the nearest neighbor method.

*2) Damage Detection:* Random rotations within a -10% to +10% range were applied to the dataset to improve the model's generalization to minor distortions, particularly those arising from extracting images from 360° video footage, as well as variations encountered in real-world scenarios. This approach enhances the model's ability to accurately detect damage across different contexts.

## III. Model Architecture and Training

### A. YOLOv11

The latest model in the YOLO series, based on the work by Redmon et al. [7], is employed for object detection, classification, and localization of buildings in complex post-disaster scenes. YOLOv11 processes the entire image in a single forward pass, enabling real-time detection capabilities crucial for rapid disaster response. The model outputs bounding boxes around detected buildings, along with category classifications and confidence scores (Figure 3).

## B. *ResNet50*

For the damage classification task, we employed a transfer learning approach by utilizing a pre-trained ResNet50 model as our base model [8]. The ResNet50 architecture, known for its deep residual learning framework, was loaded with weights pre-trained on the ImageNet dataset. We removed the top layers of the pre-trained model and added custom layers tailored to our specific classification task, including a global average pooling 2D layer (reduces the spatial dimensions of the feature maps output by the base model, creating a fixed-size vector regardless of input image size), a dense layer with 1024 units and ReLU activation (learns higher-level features from the pooled output of the base model), and an output dense layer with SoftMax activation (converts the raw output scores into probabilities that sum to 1 across all classes), where the number of units corresponded to the number of damage classes in our dataset (Figure 1). A categorical cross entropy loss function then measures how well the model performs on a given dataset (N) by comparing the predicted probability ($\hat{y}_{i,c}$) distribution over $C$ classes to the true class labels ($y_{i,c}$), as shown below:

$$Loss = -\sum_{i=1}^{N}\sum_{c=1}^{C} y_{i,c} \log(\hat{y}_{i,c})$$

# IV. Results and Discussion

The evaluation metrics of the YOLOv11 model provide valuable insights into its performance in detecting and localizing buildings within post-disaster scenes. The consistent decrease in training losses (box_loss, cls_loss, and dfl_loss) indicates that the model is learning effectively. The overall accuracy reached 60.83%, while the mean Average Precision (mAP) at 50% IoU and 50-95% IoU thresholds showed an upward trend, reaching approximately 0.45 and 0.27, respectively, by the end of training. These results reflect the inherent challenges in the dataset [2], including occlusions from surrounding objects, variations in lighting conditions, and image quality. We plan to address these challenges in future studies through large-scale testing across broader geographical regions and improved preprocessing techniques. In contrast, the ResNet50 model achieved an accuracy of 90.28%, demonstrating its strong ability to distinguish between different damage states. All training and evaluation metrics are summarized in Table 1, Figures 3 and 4.

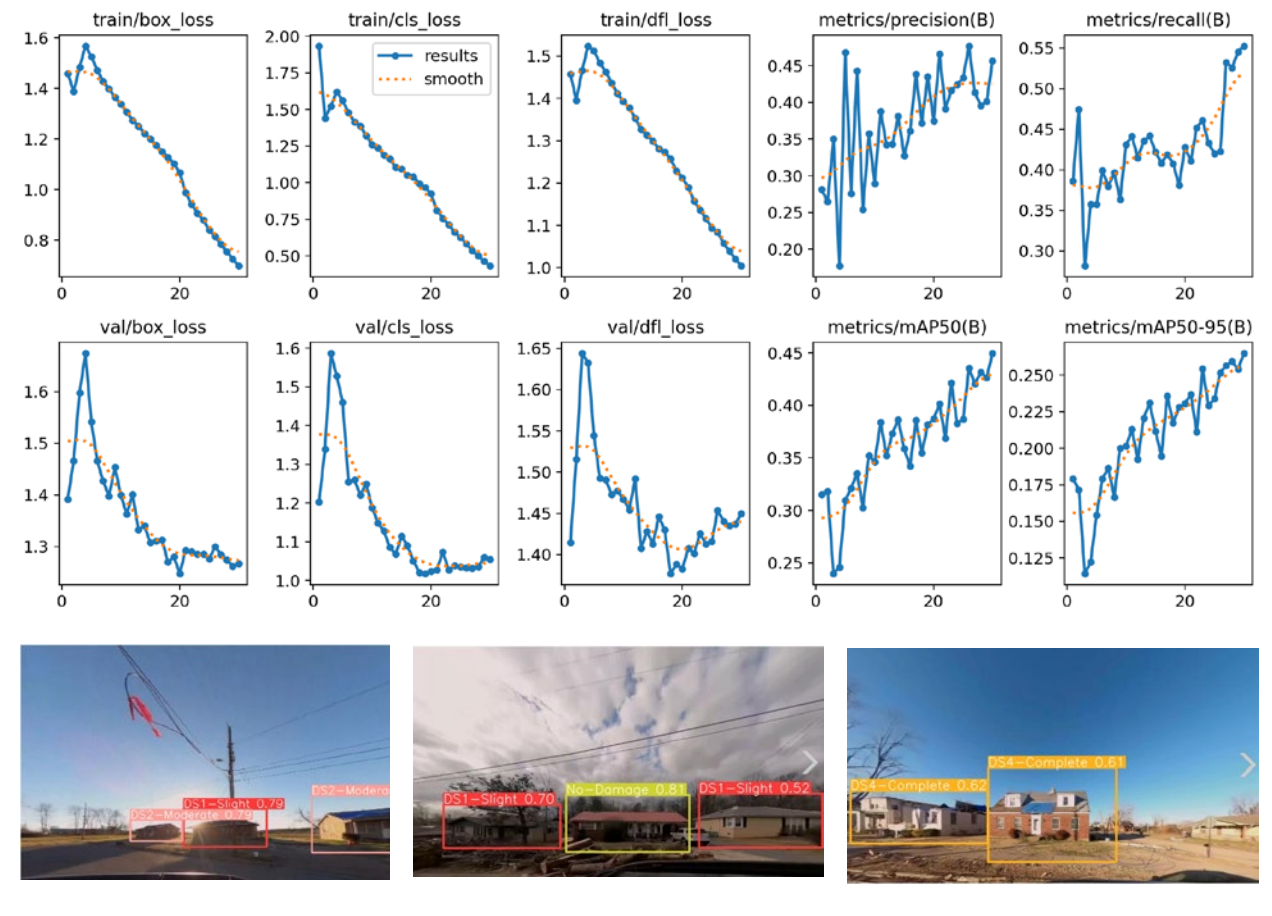

**Fig. 3.** Evaluation Metrics (YOLOv11): Training Losses, Precision, Recall, Mean Average Precision, and Sample Predictions with Confidence Scores.

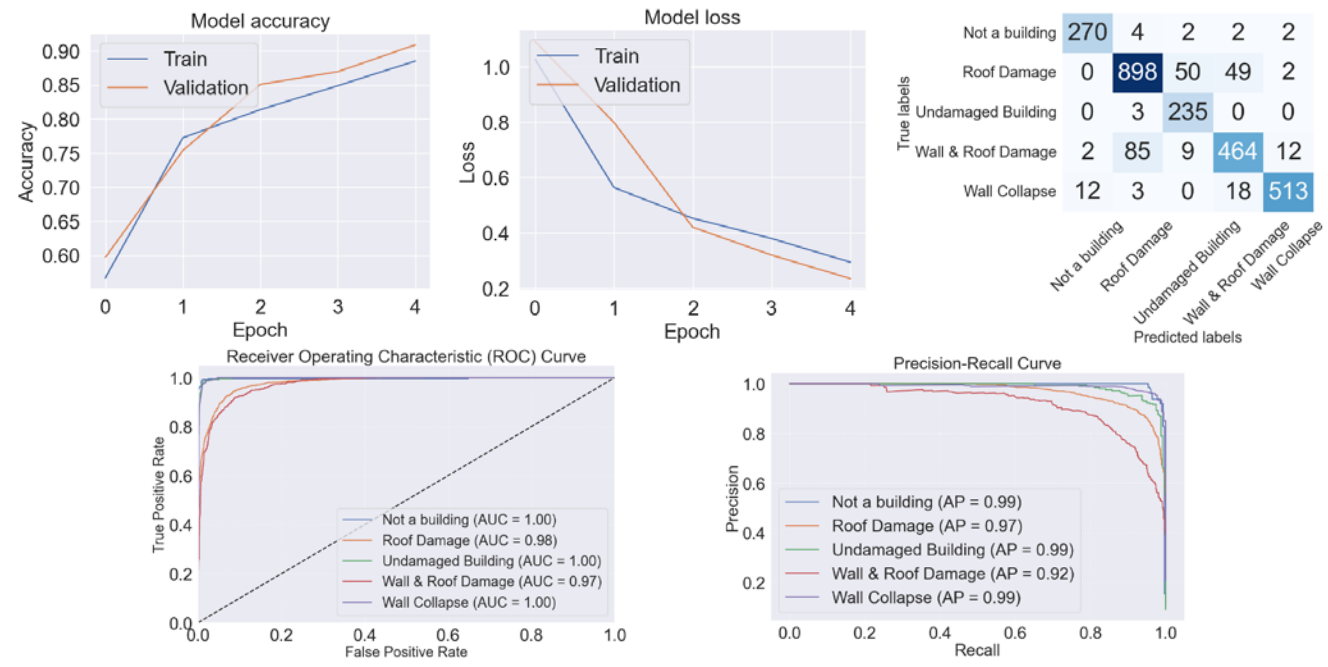

**Fig. 4.** Evaluation Metrics (ResNet50): Learning Curves (Model Accuracy and Loss), Precision-Recall Curve, ROC Curve, and the Confusion Matrix

# V. Conclusion

This research advances post-disaster building assessment through innovative deep learning solutions. Our dual-model approach demonstrates compelling results: ResNet50 achieved 90.28% accuracy in damage classification (1529ms/image), while YOLOv11 enabled rapid detection at 3ms/frame with 60.83% accuracy. This significant improvement over manual methods transforms week-long assessments into near real time analysis [2, 8], enabling faster emergency response and more efficient resource allocation. While challenges remain with class imbalance, our system provides a strong foundation for accelerated disaster response. Future work will focus on expanding testing scope and enhancing model robustness. This research marks a crucial step toward improving windstorm disaster resilience, offering communities a powerful tool that combines speed, accuracy, and standardization in post-disaster reconnaissance efforts.

# References

[1] W. “Lisa” Wang et al., “Application of Multidisciplinary Community Resilience Modeling to Reduce Disaster Risk: Building Back Better,” Journal of Performance of Constructed Facilities, vol. 38, no. 3. American Society of Civil Engineers (ASCE), Jun. 2024. doi: 10.1061/jpcfev.cfeng-4650.

[2] B. Johnston et al., “Interdisciplinary data collection for empirical community-level recovery modelling,” IABSE Reports, vol. 120. International Association for Bridge and Structural Engineering (IABSE), pp. 1260–1267, 2024. doi: 10.2749/manchester.2024.1260.

[3] J. D. Bray, J. D. Frost, E. M. Rathje, and F. E. Garcia, “Turning Disaster into Knowledge in Geotechnical Earthquake Engineering,” Geotechnical Earthquake Engineering and Soil Dynamics V. American Society of Civil Engineers, pp. 186–200, Jun. 07, 2018. doi: 10.1061/9780784481462.018.

[4] A. B. Khajwal and A. Noshadravan, “An uncertainty-aware framework for reliable disaster damage assessment via crowdsourcing,” International Journal of Disaster Risk Reduction, vol. 55. Elsevier BV, p. 102110, Mar. 2021. doi: 10.1016/j.ijdrr.2021.102110.

[5] A. S. Albahri et al., “A systematic review of trustworthy artificial intelligence applications in natural disasters,” Computers and Electrical Engineering, vol. 118. Elsevier BV, p. 109409, Sep. 2024. doi: 10.1016/j.compeleceng.2024.109409.

[6] R. Duggal et al., “Building structural analysis based Internet of Things network assisted earthquake detection,” Internet of Things, vol. 19. Elsevier BV, p. 100561, Aug. 2022. doi: 10.1016/j.iot.2022.100561.

[7] J. Redmon, S. Divvala, R. Girshick, and A. Farhadi, “You Only Look Once: Unified, Real-Time Object Detection,” 2016 IEEE Conference on Computer Vision and Pattern Recognition (CVPR). IEEE, pp. 779–788, Jun. 2016. doi: 10.1109/cvpr.2016.91.

[8] K. He, X. Zhang, S. Ren, and J. Sun, “Deep Residual Learning for Image Recognition,” 2016 IEEE Conference on Computer Vision and Pattern Recognition (CVPR). Jun. 2016. doi: 10.1109/cvpr.2016.90.

[9] P. S. Crawford, “Rapid Disaster Data Dissemination and Vulnerability Assessment through Synthesis of a Web-Based Extreme Event Viewer and Deep Learning,” Advances in Civil Engineering, vol. 2018, no. 1. Wiley, Jan. 2018. doi: 10.1155/2018/7258156.

[10] Graettinger, Andrew, et al. "Tornado damage assessment in the aftermath of the May 20th 2013 Moore Oklahoma tornado," 2014.